\documentclass[letterpaper]{article} % DO NOT CHANGE THIS
\usepackage{aaai2026}    
\usepackage{times}  % DO NOT CHANGE THIS
\usepackage{helvet}  % DO NOT CHANGE THIS
\usepackage{courier}  % DO NOT CHANGE THIS
\usepackage[hyphens]{url}  % DO NOT CHANGE THIS
\usepackage{graphicx} % DO NOT CHANGE THIS
\def\UrlFont{\rm}  % DO NOT CHANGE THIS
\usepackage{natbib}  % DO NOT CHANGE THIS AND DO NOT ADD ANY OPTIONS TO IT
\usepackage{caption} % DO NOT CHANGE THIS AND DO NOT ADD ANY OPTIONS TO IT
\usepackage{algorithm}
\usepackage{algorithmic}
\usepackage{amsmath}
\usepackage{amssymb}
\usepackage{xcolor}
\usepackage{booktabs} 
\usepackage{multirow}  
\usepackage{graphicx} 
\usepackage{bibentry}
\usepackage{newfloat}
\usepackage{listings}
\DeclareCaptionStyle{ruled}{labelfont=normalfont,labelsep=colon,strut=off}% DO NOT CHANGE THIS
\floatstyle{ruled}
\newfloat{listing}{tb}{lst}{}
\floatname{listing}{Listing}

\DeclareMathOperator*{\argmin}{arg\,min}
\nocopyright 
\title{Task-Aware Adaptive Modulation: A Replay-Free and Resource-Efficient\\Approach For Continual Graph Learning}

\author {
    Jingtao Liu\textsuperscript{\rm 1},
    Xinming Zhang\textsuperscript{\rm 1}\thanks{Corresponding author}
    }
\affiliations {
    \textsuperscript{\rm 1} University of Science and Technology of China , China \\
    jingt-liu@mail.ustc.edu.cn, xinming@ustc.edu.cn
}

\usepackage{bibentry}
\begin{document}

\maketitle
% 持续图形学习（CGL）旨在持续学习新知识而不忘记旧知识，这是现实世界图形应用的关键能力。然而，现有的方法面临着两个根本性的挑战：1)稳定性-可塑性困境：基于重放的方法通常会造成两者之间的不平衡，同时产生巨大的存储成本。2)资源繁重的预训练：领先的无重放方法严重依赖于广泛的预训练骨干，这种依赖造成了巨大的资源负担。在本文中，我们认为克服这些挑战的关键不在于重放数据或微调整个网络，而在于动态调制冻结主干的内部计算流。我们假设轻量级的、特定于任务的模块可以有效地引导GNN的推理过程。受此启发，我们提出了\textbf{任务}感知\textbf{自适应}\textbf{调制}\textbf{（TAAM），这是一种资源}高效、无重玩的方法，为解决稳定性-可塑性困境开辟了一条新途径。TAAM的核心是它的神经突触调制器（NSM），它经过训练，然后为每个任务冻结以存储专家知识。一个关键的原型引导策略管理这些调制器：1)对于训练，它通过深度复制类似的过去调制器来初始化一个新的NSM，以促进知识转移。2)对于推理，它为每个任务选择最相关的冻结NSM。这些“即插即用”的nsm插入到冻结的GNN骨干中，对其内部流进行细粒度、节点关注的调制，这与先前方法的静态扰动不同。广泛的实验表明，TAAM在六个GCIL基准数据集上全面优于最先进的方法。
\begin{abstract}

Continual Graph Learning (CGL) focuses on acquiring new knowledge while retaining previously learned information, essential for real-world graph applications. Current methods grapple with two main issues:1) The Stability-Plasticity Dilemma: Replay-based methods often create an imbalance between the Dilemma, while incurring significant storage costs.2) The Resource-Heavy Pre-training: Leading replay-free methods critically depend on extensively pre-trained backbones,this reliance imposes a substantial resource burden.In this paper, we argue that the key to overcoming these challenges lies not in replaying data or fine-tuning the entire network, but in dynamically modulating the internal computational flow of a frozen backbone. We posit that lightweight, task-specific modules can effectively steer a GNN's reasoning process.Motivated by this insight, we propose \textbf{T}ask-\textbf{A}ware \textbf{A}daptive \textbf{M}odulation (TAAM), a replay-free,resource-efficient approach that charts a new path for navigating the stability-plasticity dilemma.TAAM's core is its Neural Synapse Modulators (NSM), which are trained and then frozen for each task to store expert knowledge. A pivotal prototype-guided strategy governs these modulators: 1) For training, it initializes a new NSM by deep-copying from a similar past modulator to boost knowledge transfer. 2) For inference, it selects the most relevant frozen NSM for each task. These NSMs insert into a frozen GNN backbone to perform fine-grained, node-attentive modulation of its internal flow—different from the static perturbations of prior methods.Extensive experiments show that TAAM comprehensively outperforms state-of-the-art methods across six GCIL benchmark datasets.The code will be released upon acceptance of the paper.

\end{abstract}

% Links section - only shown in camera-ready version
% \ifdefined\aaaianonymous
% NOTE: For anonymous submissions, do not include links that could reveal your identity
%     % \link{Code}{https://aaai.org/example/code}
% \else
% % Uncomment the following to link to your code, datasets, an extended version or similar.
% % You must keep this block between (not within) the abstract and the main body of the paper.
% \fi

\begin{figure}[t]
    \centering
    \includegraphics[width=1\columnwidth]{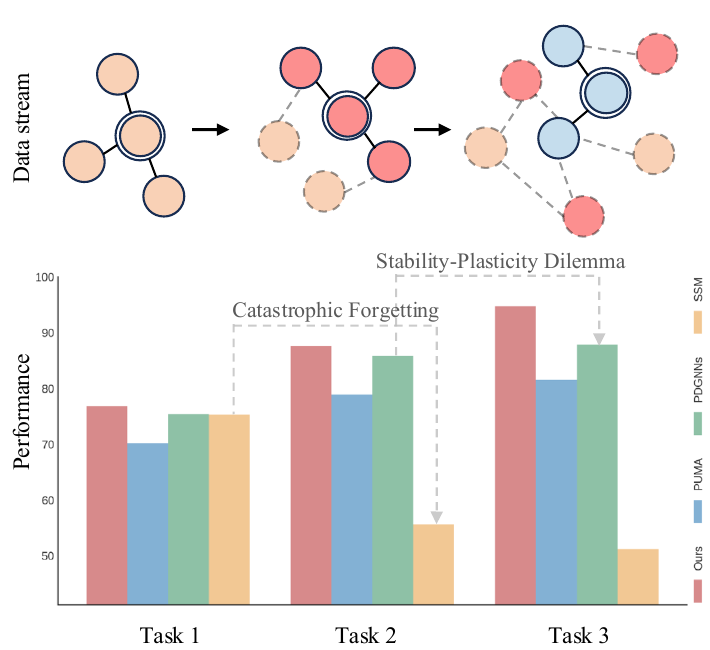}
    \caption{Top: A typical CGL scenario where a model must adapt to a stream of new tasks (indicated by new node colors) in an evolving graph. Bottom: An illustration of the Stability-Plasticity Dilemma on the Citeseer benchmark. While strong replay-based methods effectively mitigate the catastrophic forgetting of early tasks, they exhibit diminished plasticity.}
    \label{fig:single_column}
\end{figure}
% Version-specific content
% \ifdefined\aaaianonymous
\section{Introduction}

While Graph Neural Networks (GNNs) have demonstrated remarkable success on static graphs, real-world graph data is often dynamic and evolving.To address this, Continual Graph Learning (CGL) has become a key research field~\cite{Tian2024ContinualLO}, where models must learn from an evolving stream of tasks while retaining previously acquired knowledge (Figure~\ref{fig:single_column}, top). Among CGL paradigms, Graph Class-Incremental Learning (GCIL) presents a particularly challenging and practical frontier. Its core tenet is that task identifiers (IDs) are unavailable during inference, a scenario mirroring real-world applications like drug discovery, where new compound families emerge~\cite{Zhang2022CGLBBT}, or social networks, where user communities evolve over time~\cite{Wang2020LifelongGL}. This task-agnostic requirement introduces a dual challenge: first, overcoming catastrophic forgetting, and second, resolving the ``inter-task class separation'' problem---the difficulty of distinguishing between similar classes from different tasks without explicit context~\cite{Su2024TowardsRG}.

To address these challenges, researchers have primarily explored three paradigms: regularization-based, replay-based, and parameter-isolation methods~\cite{Zhang2022CGLBBT,Zhang2023ContinualLO}. Replay-based approaches, which store and rehearse historical data, can effectively mitigate forgetting but introduce significant computational and storage overhead, along with potential data privacy risks. Critically, they face a fundamental trade-off between preserving old knowledge (stability) and acquiring new information (plasticity)(Figure~\ref{fig:single_column}, Bottom).In contrast, parameter-isolation methods offer a more direct path to overcoming forgetting. By dedicating distinct parameters to different tasks, these approaches structurally prevent knowledge interference~\cite{Zhang2023ContinualLO,Zhang2021HierarchicalPN}.

%最近，提示学习（Prompt Learning）作为一种参数高效的实现方法，在计算机视觉领域取得了巨大的成功\cite{Zhang2022CGLBBT,Zhang2023ContinualLO}。这种范式很快被适应到图域\cite{Liu2023GraphPromptUP,Wang2025PromptDrivenCG}，像TPP这样的开创性工作推断任务id，用小的、可学习的“图提示”\cite{niu2024replayandforgetfree}来调节冻结的GNN。然而，这种模式的成功是由一项重大成本支撑的——“训练前税”。其有效性关键取决于一个强大的、预先训练的GNN主干，以确保知识可转移\cite{Fang2022UniversalPT,Liu2023GapformerGT}。此外，大多数图形提示方法采用单一的通用提示来调节模型的输入或输出，仅提供对GNN内部计算流的粗粒度控制，从而限制了其表达和自适应能力\cite{niu2024replayandforgetfree}。这种对预训练的依赖在图域尤为繁重。与NLP和视觉不同，缺乏标准化、通用的预训练gnn，跨异构图结构对齐数据仍然是一个巨大的挑战\cite{Hu2020OpenGB,Tang2023GraphGPTGI}。

Recently, Prompt Learning has gained prominence as a parameter-efficient implementation of this idea, achieving great success in computer vision~\cite{Zhang2022CGLBBT,Zhang2023ContinualLO}. This paradigm was quickly adapted to the graph domain~\cite{Liu2023GraphPromptUP,Wang2025PromptDrivenCG} with pioneering works like TPP inferring task IDs to condition a frozen GNN with small, learnable ``graph prompts"~\cite{niu2024replayandforgetfree}. However, the success of this paradigm is underpinned by a significant cost—a "pre-training tax." Its effectiveness critically hinges on a powerful, pre-trained GNN backbone to ensure knowledge transferability~\cite{Fang2022UniversalPT,Liu2023GapformerGT}. Furthermore, most graph prompting methods employ a single, universal prompt to modulate the model's input or output, offering only coarse-grained control over the GNN's internal computational flow and thus limiting its expressive and adaptive capacity~\cite{niu2024replayandforgetfree}. This dependency on pre-training is particularly burdensome in the graph domain. Unlike in NLP and vision, there is a lack of standardized, universal pre-trained GNNs, and aligning data across heterogeneous graph structures remains a formidable challenge~\cite{Hu2020OpenGB,Tang2023GraphGPTGI}.

This confluence of challenges motivates a fundamental research question: How can we design a CGL approach that effectively resolves the stability-plasticity dilemma without relying on data replay or an expensive, pre-trained model, thereby achieving truly resource-efficient graph continual learning?

% 在本文中，我们通过重新思考如何在资源高效、无回收的环境中管理稳定性和可塑性之间的权衡来应对这一挑战。我们的方法通过利用冻结的GNN主干并将特定任务的知识封装在轻量级调制器中来保持稳定性。我们认为，解锁可塑性的关键不在于昂贵的预训练，而在于智能地从模型自己过去的经验中回收知识。我们假设，通过建立在最相关的先验知识的基础上，可以更有效地学习新任务。
In this paper, we tackle this challenge by rethinking how the stability-plasticity trade-off can be managed in a resource-efficient, repaly-free context. Our approach maintains stability by leveraging a frozen GNN backbone and encapsulating task-specific knowledge within lightweight modulators. The key to unlocking plasticity, we argue, lies not in expensive pre-training, but in intelligently recycling knowledge from the model's own past experiences. We posit that new tasks can be learned far more effectively by building upon the foundations of the most relevant prior knowledge.

%为此，我们引入了任务感知自适应调制（TAAM），这是一种资源高效且无重复的方法，它采用轻量级神经突触调节器（NSM），即插即用模块插入到冻结的GNN中，以引导GNN的推理过程。这些nsm由一个关键的原型引导策略控制：1)对于训练，TAAM不是从一张白纸初始化每个新任务的调制器，而是使用来自最类似的过去任务的参数作为强大的“热启动”，促进有针对性的知识转移。“一旦初始化，这些调制器就会对GNN的内部计算流进行细粒度、节点关注的调制。2)对于推理，它为每个任务输入选择最相关的冻结NSM，然后由统一的线性分类器处理调制后的输出，该分类器保持冻结状态，并可以使用新的类参数进行扩展，从而防止决策层的灾难性遗忘。最终，TAAM实现了优异的稳定性-塑性平衡。我们将我们的主要贡献总结如下
To this end, we introduce Task-Aware Adaptive Modulation (TAAM), a repaly-free and resource-efficient approach that it employs lightweight Neural Synapse Modulators (NSM), these modules inserted into a frozen GNN to steer a GNN's reasoning process. These NSMs is governed by a pivotal prototype-guided strategy:For training,instead of initializing each new task's modulator from a blank slate, TAAM facilitates targeted knowledge transfer, using parameters from the most analogous past task as a powerful ``warm start."Once initialized, these modulators perform fine-grained, node-attentive modulation of the GNN's internal computational flow. For inference, it selects the most relevant frozen NSM for each task input and the modulated output are then handled by a unified linear classifier, which is kept frozen and can be extended with new class parameters, thereby preventing catastrophic forgetting at the decision layer. Ultimately, TAAM achieves a superior stability-plasticity balance.We summarize our main contributions as follows.

% 我们提出了一种新的CGL方法TAAM，它同时消除了对数据重播和模型预训练的依赖。通过一种新颖的原型引导策略来指导新任务调节器的创建和现有专家的选择，可以实现知识的前向传递而不会忘记。
%我们设计了一个动态的节点关注机制NSM。这些自适应调制器有效地封装了特定于任务的知识，与静态的通用提示技术相比，提供了更细粒度和更具表现力的GNN内部表示调制。
%我们在六个GCIL基准数据集上进行了广泛的实验。我们的研究结果表明，即使使用随机初始化的GNN骨干网，TAAM也全面且显著优于现有的最先进方法，验证了我们方法的有效性和泛化能力。
\begin{itemize}
    \item We propose TAAM, a novel CGL approach that simultaneously eliminates the reliance on data replay and model Pre-training.Through a novel prototype-guided strategy to guide both the creation of new task modulators and the selection of existing experts, the forward transmission of knowledge can be achieved without forgetting.
    \item We design NSM, a dynamic and node-attentive mechanism. These adaptive modulators effectively encapsulate task-specific knowledge, offering a more fine-grained and expressive modulation of the GNN's internal representations compared to static, general prompting techniques.
    \item We conduct extensive experiments on six GCIL benchmark datasets. Our results demonstrate that TAAM, even when using a randomly initialized GNN backbone, comprehensively and significantly outperforms existing state-of-the-art methods, validating the effectiveness and generalization capability of our approach.
\end{itemize}

\begin{figure*}[t!]
  \centering
  \includegraphics[width=0.80\linewidth]{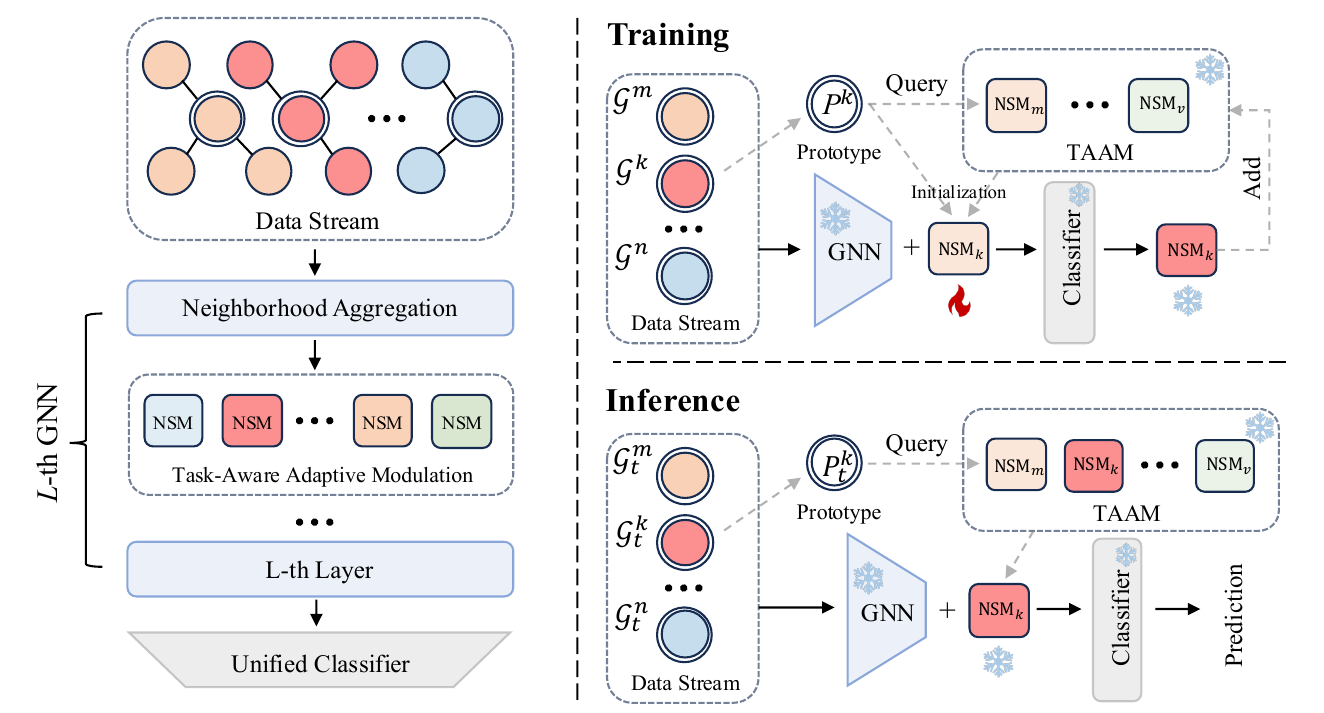}
  \caption{Overview of TAAM framework.TAAM is inserted the frozen GNN backbone,to steer a GNN's reasoning process in stream of graph data.The diagram illustrates the prototype-guided strategy of TAAM.Training Phase: For a new graph task $\mathcal{G}^k$, a corresponding prototype $P^k$ is extracted. This $P^k$ is used to query the bank of existing frozen Neural Synapse Modulators (NSM) to find the most similar past task. The parameters from the selected modulator and the $P^k$ are then used to initialize the new $\mathrm{NSM}_k$.After training is complete,  $\mathrm{NSM}_k$ frozen and added to the TAAM module bank.Inference Phase:For test task k $\mathcal{G}^k_t$ , the prototype strategy used it prototype $P^k_t$ selects the most relevant ``expert" NSM from the bank.}
  \label{fig:overstructure}
\end{figure*}

\section{Related Works}

\textbf{Replay-based Methods.} A dominant paradigm in Continual Graph Learning (CGL) is the use of replay-based methods, which maintain a memory buffer of representative exemplars from past tasks. During training on a new task, these stored exemplars are rehearsed alongside the new data, mitigating catastrophic forgetting through joint optimization~\cite{liu2023cat, liu2024puma, Zhang2022SparsifiedSubgraphMemoryforCGRL, Zhang2024Topology-awareembeddingmemoryforcl}. The sophistication of what is stored has evolved, from representative nodes~\cite{Zhou2021Overcomingcatastrophicforgetting} to more complex structures. For instance, CaT~\cite{liu2023cat} and PUMA~\cite{liu2024puma} focus on graph condensation techniques to create small, synthetic graphs for replay. Others compress topological information, either by storing sparsified computation subgraphs (SSM~\cite{Zhang2022SparsifiedSubgraphMemoryforCGRL}) or by decoupling the model from the graph via compressed topology-aware embeddings (PDGNNs~\cite{Zhang2024Topology-awareembeddingmemoryforcl}). While these methods often demonstrate strong performance, their reliance on storing and re-processing historical data introduces significant computational and memory overhead, along with potential data privacy concerns.

\textbf{Prompt Learning and Parameter-Efficient Methods.}
An alternative direction is rooted in Parameter-Efficient Transfer Learning (PETL), a paradigm where large pre-trained models are adapted to new tasks by updating only a small subset of parameters. Representative methods like Adapters achieve this by inserting lightweight modules between the layers of a backbone network to reduce fine-tuning costs~\cite{houlsby2019parameterefficienttransferlearningnlp,Gao2024BeyondPL}. 

Building on this foundation, prompt learning has emerged as a leading PETL technique. It guides a model's behavior for specific tasks by injecting learnable ``prompts,'' and is widely adopted for its excellent few-shot and transferability performance under the ``pre-train, prompt, predict'' framework~\cite{liu2021pretrainpromptpredictsystematic}. Inspired by its success, researchers have recently extended this paradigm to Graph Neural Networks (GNNs). For instance, UPT conditions a frozen GNN on structural prompts to enhance few-shot and cross-domain performance~\cite{Fang2022UniversalPT}, while GraphPrompt unifies pre-training and downstream tasks through a shared interface to improve generalization~\cite{Liu2023GraphPromptUP}. Other works like TPP infers task IDs to select the appropriate task-specific prompt for conditioning the GNN~\cite{niu2024replayandforgetfree}.

Although these methods achieve impressive parameter-efficient knowledge transfer, their effectiveness critically hinges on the existence of a powerful, resource-heavy pre-trained backbone. This ``pre-training tax'' is particularly burdensome in the graph domain, where universal, foundational GNNs are not readily available. This limitation motivates our research into a new paradigm that delivers the efficiency of parameter isolation without the prerequisite of pre-training.

\section{Preliminaries}

\noindent\textbf{Graph Neural Networks.} A GNN computes the hidden representation for a node $h_i^{(l)}$ at a given layer $l$ by aggregating features from its neighbors $\mathcal{N}(i)$. \cite{gcn} This process is generally defined by applying a transformation matrix $\mathbf{W}^{(l)}$ and a non-linear activation function $\sigma(\cdot)$:
\begin{equation}
    h_i^{(l)}=\sigma\left(\sum_{j\in\mathcal{N}(i)}\mathcal{A}_{ij}h_j^{(l-1)}\mathbf{W}^{(l)}\right)
\end{equation}
Here, $\mathcal{A}$ is a matrix defining the aggregation strategy, and $h_i^{(0)}$ is the initial node feature.

\vspace{0.5em} % Adds a little vertical space between paragraphs
SGC ~\cite{Wu2019SimplifyingGC} streamlines the GNN architecture by removing intermediate non-linear activation functions. This decouples the model into two distinct stages:

\begin{enumerate}
    \item \textit{Linear Feature Propagation:} The initial node features $\mathbf{X}$ are aggregated over a K-hop neighborhood by being multiplied by the $K$-th power of the normalized adjacency matrix $\mathbf{S}$. This creates a fixed, pre-processed feature matrix $\mathbf{X}'$:
    \begin{equation}
        \mathbf{X}' = \mathbf{S}^K \mathbf{X}
    \end{equation}

    \item \textit{Linear Classification:} A simple linear classifier with a trainable weight matrix $\mathbf{W}$ is applied to the propagated features for the final prediction:
    \begin{equation}
        \hat{\mathbf{Y}} = \text{softmax}(\mathbf{S}^K \mathbf{X} \mathbf{W})
    \end{equation}
\end{enumerate}

\textbf{GCIL Problems.}
The work of this paper is oriented towards Graph Class-Incremental Learning (GCIL), which is a more challenging setup than task-increment.The objective is, given a sequence of tasks $\{\tau_1, \dots, \tau_N\}$ with disjoint class sets $\{\mathcal{C}_1, \dots, \mathcal{C}_N\}$, to learn a unified model $f$ capable of predicting a label $y \in \bigcup_{k=1}^N \mathcal{C}_k$ for any given node, its task IDs cannot be obtained during inference.

%TAAM框架概述。TAAM在固定的GNN主干中插入，引导GNN在图数据流中的推理过程。该图说明了TAAM的原型引导策略。训练阶段：对于新的图任务$\mathcal{G}^k$，从其数据流中提取相应的原型$P^k$。该原型用于查询现有冻结的神经突触调制器（nsm）库，以找到最相似的过去任务。然后使用所选调制器和原型Pk的参数初始化新的。训练完成后，NSM_k被冻结并添加到TAAM模块库中。推理阶段：对于测试任务k，原型策略从库中选择最相关的冻结“专家”NSM。

\section{Methods}

The foundation of TAAM is a standard GNN backbone, $f_{gnn}(\cdot)$, whose parameters, $\theta_{gnn}$, are frozen to serve as a stable feature extractor. This frozen backbone acts as a stable, general-purpose feature extractor. To adapt this static model to a sequence of tasks $\mathcal{T} = \{\tau_1, \tau_2, \dots, \tau_N\}$, we introduce a set of lightweight, task-specific \textbf{Neural Synapse Modulators}(NSM), denoted as $\{\text{NSM}_1, \text{NSM}_2, \dots, \text{NSM}_N\}$,which general implementation can be seen from Figure \ref{fig:overstructure}. Architecturally, these modulators are inserted between the layers of the frozen GNN. Each $\text{NSM}_{\tau}$, with parameters $\phi_{\tau}$, is designed to steer the computational flow of the frozen GNN for a specific task $\tau$.

For a given node $v$ with initial features $\mathbf{x}_v$ from a graph $G$, the forward pass through the $L$-layer GNN is modulated at each intermediate layer $l$:
\begin{equation}
    \mathbf{h}_v^{(l+1)} = f_{gnn}^{(l)}(\mathcal{M}^{(l)}(\mathbf{h}_v^{(l)}, \phi_{\tau}))
\end{equation}
where $\mathbf{h}_v^{(l)}$ is the node representation at layer $l$, and $\mathcal{M}^{(l)}$ is the modulation function performed by $\text{NSM}_{\tau}$. The final node embeddings are passed to a unified, incrementally expanding linear classifier for prediction.

\subsection{Neural Synapse Modulator}

The NSM is the core adaptive component of TAAM. Instead of applying a single, static transformation for an entire task, the NSM generates node-specific modulation parameters, enabling node-attentive conditioning based on the features of each node.

For each GNN layer $l$ being modulated, its corresponding NSM generates a pair of modulation parameters, $[\boldsymbol{\gamma}_v^{(l)}, \boldsymbol{\beta}_v^{(l)}]$, for every node $v$. This is achieved through a node-attentive mechanism. First, a learnable task embedding, $\mathbf{e}_{\tau} \in \mathbb{R}^{d_e}$, which is unique to $\text{NSM}_{\tau}$, is linearly projected to form a set of $K$ base modulation heads, $\mathbf{B}_{\tau} \in \mathbb{R}^{K \times 2d_h}$:

\begin{equation}
    \mathbf{B}_{\tau} = \mathbf{W}_{base}\mathbf{e}_{\tau} + \mathbf{b}_{base}
\end{equation}
where $d_h$ is the hidden dimension of the node features $\mathbf{h}_v^{(l)}$. Concurrently, the input node features $\mathbf{h}_v^{(l)}$ are used to compute attention scores over these $K$ heads:
\begin{equation}
    \mathbf{a}_v = \mathbf{W}_{attn}\mathbf{h}_v^{(l)} + \mathbf{b}_{attn}
\end{equation}

The scores are normalized via a softmax function to produce attention weights, $\boldsymbol{\alpha}_v = \text{softmax}(\mathbf{a}_v) \in \mathbb{R}^{K}$. The final, node-specific modulation parameters are computed as a weighted sum of the base heads, steered by the node's own features:

\begin{equation}
    [\boldsymbol{\gamma}_v^{(l)}, \boldsymbol{\beta}_v^{(l)}] = \boldsymbol{\alpha}_v \mathbf{B}_{\tau}
\end{equation}
These parameters condition the GNN's computation via a FiLM-style affine transformation. The modulated features $\tilde{\mathbf{h}}_v^{(l)}$ are then fed into the GNN layer:
\begin{equation}
    \tilde{\mathbf{h}}_v^{(l)} = \boldsymbol{\gamma}_v^{(l)} \odot \text{LayerNorm}(\mathbf{h}_v^{(l)}) + \boldsymbol{\beta}_v^{(l)}
\end{equation}
This node-attentive mechanism allows TAAM to adapt its internal computations with much greater expressivity than static prompting techniques.

\subsection{Prototype-Guided strategy}

TAAM's ability to learn continually without pre-training or replay hinges on its prototype-guided strategy, which governs both knowledge transfer during training and knowledge retrieval during inference.

\subsubsection{Task-Aware Initialization.}

During training, when a new task $\tau_{new}$ arrives, TAAM must create a new modulator, $\text{NSM}_{new}$, without catastrophically forgetting past tasks. To foster plasticity and accelerate learning, we facilitate knowledge transfer from previously learned tasks. This is guided by task-level prototypes.

First, we compute a representative prototype $\mathbf{p}_{new}$ for the new task by averaging the initial aggregated features of its training nodes:
\begin{equation}
    \mathbf{p}_{new} = \frac{1}{|\mathcal{V}_{train}|}\sum_{v \in \mathcal{V}_{train}} f_{agg}(G_{new}, \mathbf{X}_{new})_v
\end{equation}
Next, we find the most similar past task, $\tau^*$, by comparing this new prototype to the stored prototypes of all previous tasks, $\{\mathbf{p}_1, \dots, \mathbf{p}_{new-1}\}$, using Euclidean distance as the similarity metric:
\begin{equation}
    \tau^* = argmin_{\tau' \in \{1, \dots, new-1\}} \|\mathbf{p}_{new} - \mathbf{p}_{\tau'}\|_2
\end{equation}
Instead of initializing $\text{NSM}_{new}$ from scratch, we perform a selective knowledge transfer from the most similar past task, $\text{NSM}_{\tau^*}$, to achieve a powerful ``warm start''. Specifically, we copy the structural parameters of $\text{NSM}_{\tau^*}$ (i.e., the attention and base projection weights), thereby transferring its learned modulation mechanism. 

\begin{align}
    \phi_{new}^{\text{shared}} &\leftarrow \text{copy}(\phi_{\tau^*}^{\text{shared}})
\end{align}

However, to ensure the new modulator can form a unique identity for the new task, its learnable task embedding, $\mathbf{e}_{new}$, is randomly re-initialized. This process, where the modulator's parameters $\phi_{\tau}$ are composed of shared weights $\phi_{\tau}^{\text{shared}}$ and a task embedding $\mathbf{e}_{\tau}$, is formalized as:
\begin{align}
    \mathbf{e}_{new} &\leftarrow \text{RandomInit}()
\end{align}

This hybrid initialization strategy transfers relevant inductive biases while preserving the plasticity required to learn the specifics of the new task.

During the training for task $\tau_{new}$, only the parameters $\phi_{new}$ of the newly created modulator are updated. All other modulators and the GNN backbone remain frozen. Furthermore, the final linear classifier, $f_C$, is incrementally extended. The weights corresponding to past classes are frozen, and a new output head is initialized for the classes in $\tau_{new}$, structurally preventing catastrophic forgetting at the decision layer.

\textbf{Training Objective.} To address the common issue of class imbalance within graph datasets, we employ a weighted cross-entropy loss. For the current task $\tau_{new}$, we first calculate the number of training instances $N_c$ for each class $c$. The weight for each class is then determined by its inverse frequency: $w_c = 1 / \max(N_c, 1)$. The final training objective is the weighted cross-entropy loss, computed over the training nodes $\mathcal{V}_{train}$ and the classes $C_{new}$ belonging to the new task:
\begin{equation}
\mathcal{L}_{new} = - \sum_{v \in \mathcal{V}_{train}} w_{y_v} \log\left(\frac{\exp(\mathbf{z}_{v, y'_v})}{\sum_{c' \in C_{new}} \exp(\mathbf{z}_{v, c'})}\right)
\end{equation}
where $\mathbf{z}_v$ are the logits for node $v$, and $y_v$ and $y'_v$ are its global and task-re-indexed labels, respectively.

\begin{algorithm}[htbp]
    \caption{Task-Aware Initialization}\label{alg:taam_strategy}
    \textbf{Input:} $G_{new}, \mathbf{X}_{new}, \mathcal{V}_{train},$ $\{\text{NSM}_{\tau'}\}_{\tau'=1}^{new-1}$, $\{\mathbf{p}_{\tau'}\}_{\tau'=1}^{new-1}$\\
    \textbf{Output:} trained $\text{NSM}_{new}$.
    \begin{algorithmic}[1]
        \STATE Compute prototype $\mathbf{p}_{new}$ for the new task.
        \IF{it is the first task ($new = 1$)}
            \STATE Initialize a new modulator $\text{NSM}_{new}$ with random parameters.
        \ELSE
            \STATE Find the most similar past task $\tau^*$ by comparing $\mathbf{p}_{new}$ with all stored prototypes $\{\mathbf{p}_{\tau'}\}$.
            \STATE Initialize structural parameters of $\text{NSM}_{new}$ by copying from $\text{NSM}_{\tau^*}$. \COMMENT{Transfer knowledge}
            \STATE Randomly re-initialize the task embedding $\mathbf{e}_{new}$ of $\text{NSM}_{new}$. \COMMENT{Adapt to new task}
        \ENDIF
        \STATE Freeze the GNN backbone and all previously trained modulators $\{\text{NSM}_{\tau'}\}_{\tau'=1}^{new-1}$.
        \FOR{each training epoch $e=1$ to $E$}
            \STATE Obtain logits $\mathbf{Z}$ via a forward pass on $G_{new}$ using the current $\text{NSM}_{new}$.
            \STATE Compute the class-balanced cross-entropy loss $\mathcal{L}_{new}$.
            \STATE Update only the parameters $\phi_{new}$ of $\text{NSM}_{new}$ via backpropagation.
        \ENDFOR
        \STATE Store the trained $\text{NSM}_{new}$ and its prototype $\mathbf{p}_{new}$.
        \STATE Return trained $\text{NSM}_{new}$.\\
    \end{algorithmic}
\end{algorithm}

\subsubsection{Task-Aware Retrieval.}
During inference, the task ID for a given input graph is unknown. TAAM infers the most likely task ID, $\hat{\tau}$, by applying the same prototype-matching logic. A prototype is computed for the test batch, and its distance is measured against all stored task prototypes. The ID of the closest prototype is selected as the inferred task ID:
\begin{equation}
    \hat{\tau} = \argmin_{\tau' \in \{1, \dots, N\}} \|\mathbf{p}_{test} - \mathbf{p}_{\tau'}\|_2
\end{equation}
The forward pass is then executed using the corresponding frozen modulator, $\text{NSM}_{\hat{\tau}}$, ensuring that the correct specialized knowledge is retrieved and applied for the prediction. This entire process allows TAAM to dynamically adapt to a sequence of tasks in a scalable and parameter-efficient manner.

\section{Experiments}
\begin{table}[!t]
\centering
\caption{The detailed statistics of datasets}
\label{tab:dataset_stats_transposed}
\begin{tabular}{lrrrr}
\toprule
\textbf{Dataset} & \textbf{Nodes} & \textbf{Edges} & \textbf{Classes} & \textbf{Tasks} \\
\midrule
Cora & 2,708 & 5,429 & 7 & 3 \\
Citeseer & 3,327 & 4,732 & 6 & 3 \\
CoraFull & 19,793 & 130,622 & 70 & 35 \\
Arxiv & 169,343 & 1,166,243 & 40 & 20 \\
Reddit & 232,965 & 114,615,892 & 40 & 20 \\
Products & 2,449,029 & 61,859,036 & 47 & 23 \\
\bottomrule
\end{tabular}
\end{table}

\subsubsection{Dateset.}
Following the GCL benchmark~\cite{Zhang2022CGLBBT}, six large public graph datasets are employed, including Cora~\cite{Yang2016RevisitingSL}, Citeseer~\cite{Yang2016RevisitingSL},CoraFull~\cite{McCallum2000AutomatingTC}, Arxiv ~\cite{Hu2020OpenGB}, Reddit~\cite{Hamilton2017InductiveRL} and Products\cite{Hu2020OpenGB}. The comprehensive descriptions are delineated in Table \ref{tab:dataset_stats_transposed}. For all datasets, each task is set to contain only two classes. Besides, for each class, the proportions of training, validation, and testing are set to be 0.6, 0.2, and 0.2 respectively.

\subsubsection{Implementation Details.}
We employ a 2-layer Simplifying Graph Convolutional Network (SGC) ~\cite{Wu2019SimplifyingGC} as our feature extractor. Its parameters are randomly initialized and then remain frozen.For each NSM, the dimension of the learnable task embedding, $\mathbf{e}_{\tau}$, is set to 64. The internal node-attentive mechanism uses $K=3$ heads to steer for each node.For each new task, only the parameters of the newly created NSM are trainable. 

\subsubsection{Evaluation Metrics.}
We evaluate performance using Average Accuracy (AA) and Average Forgetting (AF), where a higher AA indicates better overall accuracy and a lower AF indicates less catastrophic forgetting.

% For this table to compile, you must have the following packages
% in your document's preamble:
% \usepackage{booktabs}
% \usepackage{multirow}

\begin{table*}[!t]
\centering
\caption{Comparison of different models on different data sets in GCIL setting. The bold results are the best performance excluding Joint and Oracle. '$\uparrow$' denotes the greater value represents greater performance. '$\downarrow$' the opposite. Joint can get access to the data of all tasks, Oracle can get access to the data of all tasks and task IDs.}
\label{tab:comparison}
\resizebox{\textwidth}{!}{%
\begin{tabular}{llcccccccccccc}
\toprule
\multicolumn{1}{c}{\multirow{2}{*}{Category}} & \multicolumn{1}{c}{\multirow{2}{*}{Model}} & \multicolumn{2}{c}{Cora} & \multicolumn{2}{c}{Citeseer} & \multicolumn{2}{c}{CoraFull} & \multicolumn{2}{c}{Arxiv} & \multicolumn{2}{c}{Reddit} & \multicolumn{2}{c}{Products} \\ \cmidrule(lr){3-4} \cmidrule(lr){5-6} \cmidrule(lr){7-8} \cmidrule(lr){9-10} \cmidrule(lr){11-12} \cmidrule(lr){13-14}
\multicolumn{1}{c}{} & \multicolumn{1}{c}{} & AA(\%)($\uparrow$) & AF(\%)($\downarrow$) & AA(\%)($\uparrow$) & AF(\%)($\downarrow$) & AA(\%)($\uparrow$) & AF(\%)($\downarrow$) & AA(\%)($\uparrow$) & AF(\%)($\downarrow$) & AA(\%)($\uparrow$) & AF(\%)($\downarrow$) & AA(\%)($\uparrow$) & AF(\%)($\downarrow$) \\ \midrule
Lower bound & Finetuning & 32.3$\pm$0.2 & 95.0$\pm$0.1 & 32.3$\pm$0.2 & 80.5$\pm$0.2 & 3.5$\pm$0.5 & 95.2$\pm$0.5 & 4.9$\pm$0.0 & 89.7$\pm$0.4 & 4.9$\pm$1.2 & 97.9$\pm$3.3 & 4.6$\pm$0.7 & 88.7$\pm$0.8 \\ \midrule
\multirow{4}{*}{Regularisation} & EWC (2017) & 32.3$\pm$0.2 & 95.0$\pm$0.1 & 32.3$\pm$0.2 & 80.4$\pm$0.1 & 13.2$\pm$0.3 & 95.0$\pm$0.2 & 7.9$\pm$1.4 & 81.2$\pm$1.3 & 5.1$\pm$0.0 & 95.3$\pm$2.2 & 8.24$\pm$0.7 & 90.32$\pm$1.1 \\
& LwF(2017) & 32.2$\pm$0.2 & 94.8$\pm$0.1 & 32.7$\pm$0.0 & 80.1$\pm$0.3 & 4.0$\pm$0.6 & 93.1$\pm$0.8 & 5.9$\pm$0.1 & 90.1$\pm$0.2 & 7.0$\pm$1.0 & 97.0$\pm$1.1 & 4.92$\pm$0.4 & 50.19$\pm$0.3 \\
& MAS(2018) & 32.2$\pm$0.2 & 94.8$\pm$0.3 & 30.8$\pm$0.2 & 78.5$\pm$0.4 & 11.3$\pm$0.2 & 95.0$\pm$0.4 & 4.9$\pm$0.3 & 89.3$\pm$0.0 & 4.9$\pm$0.0 & 89.3$\pm$0.9 & 3.1$\pm$0.8 & 89.5$\pm$0.9 \\
& TWP(2021) & 32.5$\pm$0.1 & 94.7$\pm$0.2 & 32.9$\pm$0.2 & 65.5$\pm$0.7 & 19.8$\pm$3.9 & 73.3$\pm$4.2 & 4.9$\pm$0.4 & 89.8$\pm$0.4 & 6.9$\pm$0.4 & 97.1$\pm$0.4 & 3.0$\pm$0.7 & 89.7$\pm$1.0 \\ \midrule
\multirow{4}{*}{Replay} & ER-GN(2021) & 86.2$\pm$1.6 & 4.5$\pm$1.9 & 73.5$\pm$0.4 & 9.1$\pm$1.4 & 28.5$\pm$0.2 & 15.4$\pm$0.2 & 46.8$\pm$1.1 & 16.4$\pm$1.3 & 73.9$\pm$3.3 & 5.7$\pm$2.5 & 25.3$\pm$1.9 & 65.6$\pm$1.9 \\
& SSM(2022) & 88.6$\pm$0.5 & 2.7$\pm$1.0 & 63.8$\pm$2.5 & 5.2$\pm$2.1 & 75.8$\pm$0.5 & 7.1$\pm$0.5 & 47.2$\pm$0.5 & 11.6$\pm$1.5 & 94.2$\pm$0.2 & 1.4$\pm$0.1 & 62.7$\pm$0.1 & 9.4$\pm$1.7 \\
& PUMA(2023) & 88.2$\pm$1.0 & 5.7$\pm$0.5 & 70.9$\pm$0.5 & 9.8$\pm$1.4 & 65.1$\pm$0.3 & 6.9$\pm$0.7 & 62.0$\pm$0.3 & 11.0$\pm$0.4 & 96.9$\pm$0.0 & 0.7$\pm$0.0 & 73.5$\pm$0.4 & 5.3$\pm$0.3 \\
& PDGNNs(2024) & 93.9$\pm$0.1 & -2.8$\pm$0.3 & 79.9$\pm$0.5 & 5.0$\pm$0.3 & 81.2$\pm$0.1 & 3.5$\pm$0.1 & 53.2$\pm$0.3 & 12.7$\pm$0.2 & 94.8$\pm$0.0 & 3.2$\pm$0.0 & 72.3$\pm$0.2 & 11.3$\pm$0.5 \\ \midrule
\multirow{2}{*}{Parameter isolation} & TPP(2024) & 97.5$\pm$0.7 & 0.0$\pm$0.0 & 80.4$\pm$0.9 & 0.0$\pm$0.0 & 93.4$\pm$0.2 & 0.0$\pm$0.0 & 85.4$\pm$0.3 & 0.0$\pm$0.0 & 99.5$\pm$0.0 & 0.0$\pm$0.0 & 94.0$\pm$0.5 & 0.0$\pm$0.0 \\
& TAAM (Ours) & \textbf{97.6$\pm$0.4} & \textbf{0.0$\pm$0.0} & \textbf{85.9$\pm$0.3} & \textbf{0.0$\pm$0.0} & \textbf{95.0$\pm$0.5} & \textbf{0.0$\pm$0.0} & \textbf{90.4$\pm$0.2} & \textbf{0.0$\pm$0.0} & \textbf{99.5$\pm$0.0} & \textbf{0.0$\pm$0.0} & \textbf{96.2$\pm$0.3} & \textbf{0.0$\pm$0.0} \\ \midrule
\multirow{2}{*}{Upper bound} & Joint & 94.2$\pm$0.0 & - & 79.8$\pm$0.3 & - & 81.2$\pm$0.4 & - & 51.3$\pm$0.5 & - & 97.1$\pm$0.1 & - & 71.5$\pm$0.1 & - \\
& Oracle & 98.1$\pm$0.1 & - & 86.1$\pm$0.2 & - & 95.3$\pm$0.3 & - & 91.2$\pm$0.2 & - & 99.5$\pm$0.1 & - & 96.5$\pm$0.4 & - \\ \bottomrule
\end{tabular}%
}
\end{table*}

% 请在您的 LaTeX 文档导言区添加以下宏包
% \usepackage{booktabs}
% \usepackage{multirow}

\begin{table*}[t]
\centering
\caption{Performance comparison in the GCIL setting under different task division protocols.The best results, excluding the Oracle upper bound, are highlighted in \textbf{bold}. "OOM" denotes an Out-of-Memory error.}
\label{tab:gcil_results}
\resizebox{\textwidth}{!}{%
\begin{tabular}{lcccccccccccc}
\toprule
\multirow{2}{*}{Model} & \multicolumn{6}{c}{CoraFull} & \multicolumn{6}{c}{Arxiv} \\
\cmidrule(lr){2-7} \cmidrule(lr){8-13}
& \multicolumn{2}{c}{Unequally} & \multicolumn{2}{c}{Equally(10)} & \multicolumn{2}{c}{Equally(5)} & \multicolumn{2}{c}{Unequally} & \multicolumn{2}{c}{Equally(10)} & \multicolumn{2}{c}{Equally(5)} \\
& AA\%($\uparrow$) & AF\%($\downarrow$) & AA\%($\uparrow$) & AF\%($\downarrow$) & AA\%($\uparrow$) & AF\%($\downarrow$) & AA\%($\uparrow$) & AF\%($\downarrow$) & AA\%($\uparrow$) & AF\%($\downarrow$) & AA\%($\uparrow$) & AF\%($\downarrow$) \\
\midrule
SSM & 69.1$\pm$0.1 & 2.8$\pm$0.2 & 70.5$\pm$0.1 & 4.2$\pm$0.3 & 62.2$\pm$0.2 & 5.8$\pm$0.2 & 41.5$\pm$0.4 & 7.3$\pm$0.1 & 37.4$\pm$0.5 & 7.5$\pm$0.4 & 41.7$\pm$0.5 & 11.3$\pm$0.5 \\
PDGNNs & 76.8$\pm$0.1 & 2.4$\pm$0.2 & 81.3$\pm$0.1 & 3.6$\pm$0.1 & 70.4$\pm$0.0 & 4.9$\pm$0.0 & 49.8$\pm$0.4 & 8.8$\pm$0.3 & 50.2$\pm$0.3 & 10.7$\pm$0.2 & 53.1$\pm$0.5 & 12.3$\pm$0.2 \\
TPP & OOM & - & 75.6$\pm$0.4 & 0.0$\pm$0.0 & 77.3$\pm$1.0 & 0.0$\pm$0.0 & OOM & - & OOM & - & 63.4$\pm$1.3 & 0.0$\pm$0.0 \\
\textbf{TAAM} & \textbf{82.3$\pm$0.5} & \textbf{0.0$\pm$0.0} & \textbf{84.0$\pm$0.3} & \textbf{0.0$\pm$0.0} & \textbf{90.0$\pm$0.3} & \textbf{0.0$\pm$0.0} & \textbf{77.2$\pm$0.4} & \textbf{0.0$\pm$0.0} & \textbf{64.1$\pm$0.6} & \textbf{0.0$\pm$0.0} & \textbf{79.6$\pm$0.2} & \textbf{0.0$\pm$0.0} \\
\midrule
Oracle & 84.1$\pm$0.1 & - & 85.6$\pm$0.1 & - & 91.4$\pm$0.1 & - & 79.7$\pm$0.3 & - & 70.9$\pm$0.2 & - & 82.3$\pm$0.1 & - \\
\bottomrule
\end{tabular}
}
\end{table*}

\subsection{Main results}

\textbf{Performance Comparison}. As shown in Table \ref{tab:comparison}, our proposed method, \textbf{TAAM}, establishes a new state-of-the-art across six diverse GCIL benchmarks. Notably, TAAM significantly outperforms the strongest replay-based competitors on large-scale datasets like Reddit (99.5\% vs. 96.9\%) and Products (96.2\% vs. 73.5\%). Crucially, this is achieved without the substantial memory overhead, computational cost, and potential privacy risks inherent to replay-based approaches. Within the replay-free category, TAAM also demonstrates a clear advantage over TPP, the previous state-of-the-art, with a notable improvement of 4.0\% and 7.3\% in average accuracy on the Arxiv and Citeseer datasets, respectively. This performance gap stems from our method's ability to operate effectively on randomly initialized backbones, eliminating the pre-training dependency required by methods like TPP.

A standout result is TAAM's perfect management of catastrophic forgetting, achieving no forgetting on all datasets. This is a direct consequence of its architectural design, where task-specific knowledge is securely encapsulated within dedicated, frozen Neural Sub-Modules (NSMs). In contrast, regularization and even top-performing replay methods inevitably exhibit knowledge degradation (e.g., PDGNNs shows 12.7\% forgetting on Arxiv). By virtually eliminating forgetting while simultaneously maximizing accuracy on new tasks, TAAM presents a more effective solution to the \textbf{stability-plasticity trade-off}.

\begin{figure*}[t!]
    \centering
    \includegraphics[width=1\linewidth]{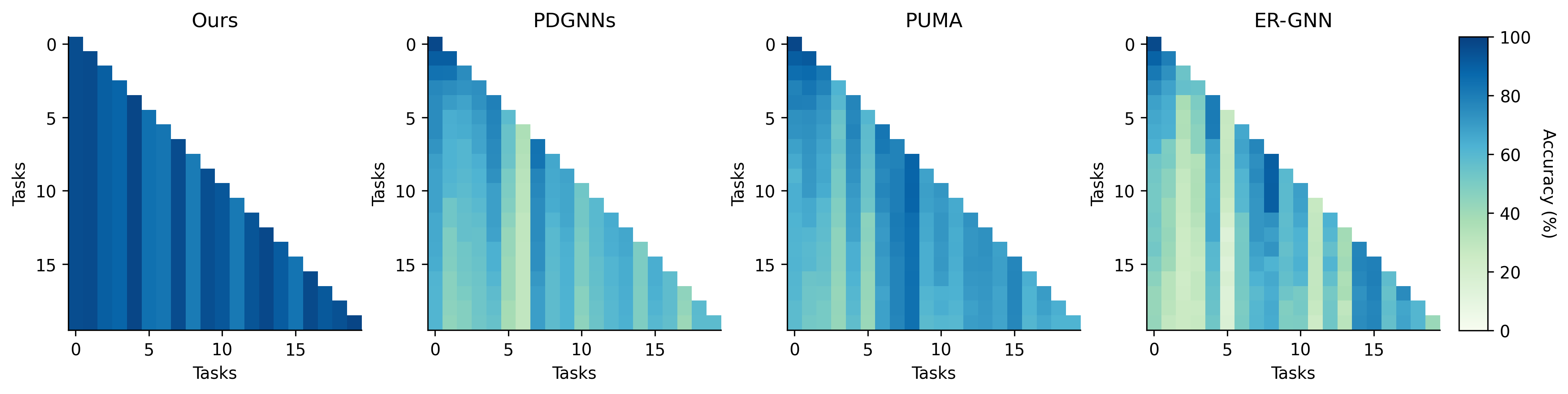}
      \caption{Performance matrices of differernt method on Arxiv dataset}
  \label{overstructure}
\end{figure*}

\textbf{Visualizing Knowledge Retention}. As depicted in Figure \ref{overstructure}, the heatmaps vividly illustrate our method's superior knowledge retention. The heatmap for Ours method exhibits a consistently dark and uniform color across the lower triangle, signifying both excellent plasticity (a dark diagonal) and near-perfect stability, as there is no color degradation down each column. In stark contrast, the heatmaps for PDGNNs, PUMA, and ER-GNN reveal a clear pattern of knowledge decay. While their diagonals indicate an ability to learn new tasks, performance on past tasks clearly degrades, evidenced by the pronounced color fading down each column.

\textbf{Analysis of Task Division Protocols}. To comprehensively evaluate the robustness of our model under various incremental learning scenarios, we designed multiple task division protocols. As shown in Table \ref{tab:gcil_results}, our experiments cover three settings: 1) Unequally: The model first learns from a large base task (20 classes on CoraFull, 10 on Arxiv) and subsequently learns from smaller incremental tasks (10 and 5 classes, respectively). 2) Equally(10) and 3) Equally(5): Each task contains a fixed number of 10 or 5 classes. Notably, these settings are more challenging than the standard 2-classes-per-task protocol used in other experiments throughout this paper.

Table \ref{tab:gcil_results} demonstrate TAAM's consistent and superior performance. The comparison with the leading replay-free baseline, TPP, is particularly revealing. TPP fails with OOM errors on the more demanding benchmarks, such as the "Unequally" split. . As indicated in the table, TPP encounters OOM errors on the more challenging benchmarks, particularly under the 'Unequally' split which features a large base task. This is attributed to its resource-intensive pre-training phase on the initial task's data, a requirement that is difficult to manage even with mini-batch strategies due to its substantial computational and memory overhead. In stark contrast, TAAM’s pre-training-free and resource-efficient architecture operates successfully across all settings. This validates its practical advantages over methods burdened by high computational demands.

\begin{table*}[!t]
\centering
\caption{Ablation results of TAAM and its variants, excluding Cora and Reddit datasets for brevity.}
\label{tab:ablation_subset}
\small % 1. 使用小一号的字体
\setlength{\tabcolsep}{3.5pt}
\begin{tabular}{cccccccccccc}
\toprule
% 保留了表头换行以优化显示效果
\multicolumn{1}{c}{\multirow{2}{*}{NSM}} & \multicolumn{1}{c}{\multirow{2}{*}{\begin{tabular}[c]{@{}c@{}}Task-Aware\\ Retrieval\end{tabular}}} & \multicolumn{1}{c}{\multirow{2}{*}{\begin{tabular}[c]{@{}c@{}}Task-Aware\\ Initialization\end{tabular}}} & \multicolumn{2}{c}{Citeseer} & \multicolumn{2}{c}{CoraFull} & \multicolumn{2}{c}{Arxiv} & \multicolumn{2}{c}{Products} \\ 
\cmidrule(lr){4-5} \cmidrule(lr){6-7} \cmidrule(lr){8-9} \cmidrule(lr){10-11}
\multicolumn{1}{c}{} & \multicolumn{1}{c}{} & \multicolumn{1}{c}{} & AA(\%)($\uparrow$) & AF(\%)($\downarrow$) & AA(\%)($\uparrow$) & AF(\%)($\downarrow$) & AA(\%)($\uparrow$) & AF(\%)($\downarrow$) & AA(\%)($\uparrow$) & AF(\%)($\downarrow$) \\ \midrule

$\checkmark$ & $\times$ & $\times$ & 31.2$\pm$0.2 & 82.0$\pm$1.0 & 2.8$\pm$0.3 & 95.3$\pm$0.3 & 4.8$\pm$0.1 & 90.7$\pm$0.4 & 3.6$\pm$1.0 & 95.7$\pm$0.6 \\

$\checkmark$ & $\checkmark$ & $\times$ & 85.5$\pm$0.4 & 0.0$\pm$0.0 & 94.7$\pm$0.4 & 0.0$\pm$0.0 & 90.1$\pm$0.3 & 0.0$\pm$0.0 & 94.9$\pm$1.1 & 0.0$\pm$0.0 \\

$\checkmark$ & $\checkmark$ & $\checkmark$ & 85.9$\pm$0.3 & 0.0$\pm$0.0 & 95.0$\pm$0.5 & 0.0$\pm$0.0 & 90.4$\pm$0.2 & 0.0$\pm$0.0 & 95.5$\pm$0.9 & 0.0$\pm$0.0 \\ \bottomrule
\end{tabular}
\end{table*}

\textbf{Efficiency Analysis}. As shown in Figure 4, the top-left quadrant of the plot represents the optimal profile of high accuracy and low computational cost. TAAM is clearly positioned in this quadrant, achieving both the highest average accuracy ($\approx$95\%) and the fastest execution time. In comparison, while TPP is competitive in accuracy, it is substantially slower. The replay-based methods, including PDGNNs, PUMA, and SSM, are not only less accurate but also significantly more computationally expensive.

\begin{figure}[t]
    \centering
    \includegraphics[width=0.9\columnwidth]{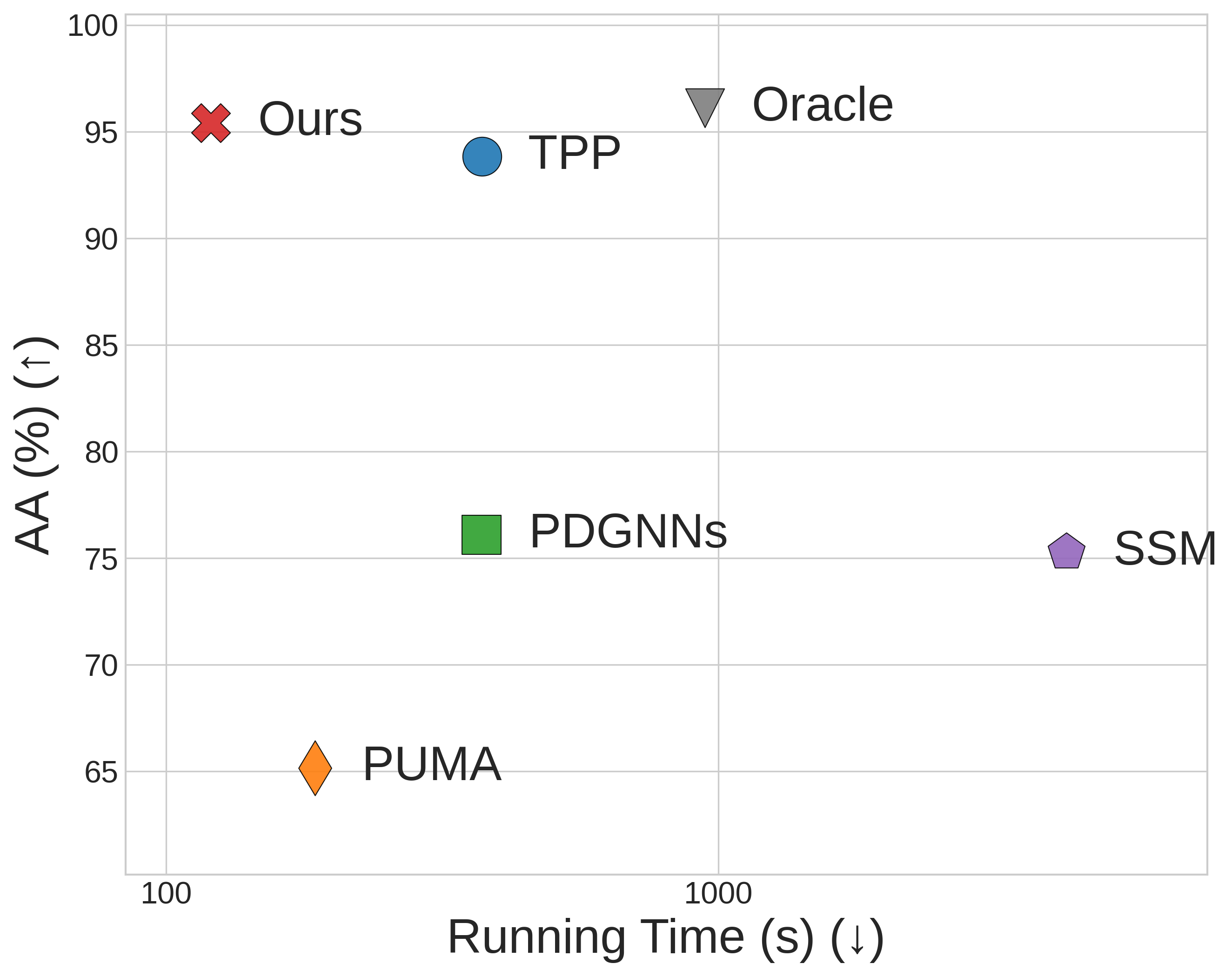}
    \caption{Comparison of running time and AA on CoraFull}
    \label{fig:single_column1}
\end{figure}

\subsection{Ablation Study}
To validate the effectiveness of our proposed components, we conducted an ablation study on TAAM, with the results presented in Table \ref{tab:ablation_subset}.The first variant, which employs only the basic NSM adapters without either strategy, performs poorly, suffering from extremely low AA and catastrophic forgetting across all datasets. The introduction of Task-Aware Retrieval yields a dramatic performance leap. For instance, on Products, AA skyrockets from 3.6\% to 94.9\%, while AF plummets from 95.7\% to a perfect 0.0\%. This demonstrates that the ability to correctly identify and activate the relevant frozen modulator at inference time is fundamental to achieving stability and preventing knowledge interference.Building on this stable foundation, the addition of Task-Aware Initialization (third row, the full TAAM model) provides a further, consistent boost in accuracy across all benchmarks. On datasets like Arxiv and Products, the AA increases from 90.1\% to 90.4\% and 94.9\% to 95.5\%, respectively. This highlights the crucial role of our initialization strategy in enhancing model plasticity. While retrieval secures past knowledge, the intelligent initialization by copying parameters from a similar past task enables the model to learn new concepts more effectively. Together,The empirical results verify that combining all the components can well balance the stability and plasticity of the model.

More supplementary experiments and details can be found in the Appendix.

\section{Conclusion} In this paper, we propose a novel continual graph learning method, TAAM, that  without relying on data replay or pre-training.  It uses a prototype-guided strategy to dynamically initialize and select lightweight, task-aware modules that modulate a frozen GNN. Extensive experiments on benchmark datasets demonstrate this approach effectively prevents catastrophic forgetting while ensuring fine-grained, efficient knowledge transfer.Future work will explore more challenging scene Settings.

\bibliography{aaai2026} 

% % --- Appendix starts here ---
\newpage % Or \clearpage
\appendix

\section{Appendix}

In this supplementary material, we provide additional details regarding our experimental setup, methodology, and results. The appendix is organized as follows:

\begin{itemize}
    \item In Section A, we detail the implementation specifics, including the hardware and software configurations.
    
    \item In Section B, we present the detailed algorithm of our proposed method, providing pseudocode for the Neural Synapse Modulator and the Task ID Retrieval processes.
    
    \item In Section C, we describe the suite of six benchmark datasets used for evaluation.
    
    \item In Section D, we elaborate on the baseline methods, categorized into replay-base, regularization, and parameter isolation approaches, against which our model is compared.
    
    \item In Section E, we formally define the evaluation metrics, Average Accuracy and Average Forgetting, based on the accuracy matrix $M \in \mathbb{R}^{T \times T}$.
    
    \item In Section F, we provide more extensive experimental results, including detailed performance comparison tables and a comprehensive hyperparameter analysis.
\end{itemize}

\section{A. Implementation Details}

All experiments are conducted on a server equipped with an Intel(R) Xeon(R) Platinum 8368Q CPU @ 2.60GHz, 256 GB of RAM, and a single NVIDIA L40 GPU. The software stack includes Python 3.8.16, PyTorch 1.10.1, DGL 0.9.0, OGB 1.3.6, and CUDA 12.4.

Our experimental setup follows the CGL benchmark~\cite{Zhang2022CGLBBT} to ensure a fair and rigorous evaluation. For all methods, we employ a unified two-layer GNN backbone with a 256-dimensional hidden layer, configured without batch normalization, dropout, or bias. We used the Adam optimizer for 200 epochs, with a learning rate of $0.005$ and a weight decay of $5 \times 10^{-4}$. Hyperparameters for all baselines are set according to their original publications, and the same random seeds are used across all runs for reproducibility.

\section{B. Algorithm}

The Neural Synapse Modulator and Task ID Retrieval processes of the proposed method are summarized in below, respectively.

\begin{algorithm}[htbp]
\caption{Neural Synapse Modulator}
\label{alg:nsm_corrected}
    \textbf{Input:} Node features $\mathbf{h}_v \in \mathbb{R}^{d_h}$, Task embedding $\mathbf{e}_{\tau} \in \mathbb{R}^{d_e}$.
    \textbf{Parameters:} Base generator weights $\mathbf{W}_{base}, \mathbf{b}_{base}$, Attention network weights $\mathbf{W}_{attn}, \mathbf{b}_{attn}$.
    
    \textbf{Output:} Parameters $(\boldsymbol{\gamma}_v, \boldsymbol{\beta}_v)$.
    \begin{algorithmic}[1]
        \STATE \COMMENT{Generate K base modulation heads from the task embedding}
        \STATE $\mathbf{B}_{\tau} \gets \mathbf{W}_{base}\mathbf{e}_{\tau} + \mathbf{b}_{base}$ \COMMENT{Eq. (5)}
        \STATE \COMMENT{Compute attention scores based on node features}
        \STATE $\mathbf{a}_v \gets \mathbf{W}_{attn}\mathbf{h}_v + \mathbf{b}_{attn} \in \mathbb{R}^{K}$ \COMMENT{Eq. (6)}
        \STATE $\boldsymbol{\alpha}_v \gets \text{Softmax}(\mathbf{a}_v) \in \mathbb{R}^{K}$
        \STATE $[\boldsymbol{\gamma}_v, \boldsymbol{\beta}_v] \gets \boldsymbol{\alpha}_v \mathbf{B}_{\tau}$ \COMMENT{Eq. (7)}
        \STATE Return $(\boldsymbol{\gamma}_v, \boldsymbol{\beta}_v)$
    \end{algorithmic}
\end{algorithm}

\begin{algorithm}[htbp]
\caption{Prototype-Based Task ID Retrieval}
\label{alg:retrieval_augmented}
    \textbf{Input:} Graph $\mathcal{G}$, Node features $\mathbf{X}$, Test node IDs $\mathcal{V}_{test}$.
    \textbf{Given:} Stored task prototypes $\mathcal{P} = \{\mathbf{p}_1, \dots, \mathbf{p}_N\}$; Aggregation function $f_{agg}$.
    
    \textbf{Output:} Inferred task ID $\hat{\tau}$.
    \begin{algorithmic}[1]
        \STATE \COMMENT{Step 1: Compute prototype for the test batch}
        \STATE $\mathbf{p}_{test} \gets \frac{1}{|\mathcal{V}_{test}|}\sum_{v \in \mathcal{V}_{test}} f_{agg}(\mathcal{G}, \mathbf{X})_v$ \COMMENT{Analogous to Eq. (9)}
        \STATE \COMMENT{Step 2: Find the closest prototype in the stored library}
        \STATE $\hat{\tau} \gets \argmin_{\tau' \in \{1, \dots, N\}} \|\mathbf{p}_{test} - \mathbf{p}_{\tau'}\|_2$ \COMMENT{Eq. (10)}
        \STATE Return $\hat{\tau}$
    \end{algorithmic}
\end{algorithm}

\section{C. Details of Datesets}
We evaluate our method on a suite of six benchmark datasets commonly used in graph machine learning and Continual Graph Learning (CGL). 

\begin{itemize}
    \item \textbf{Cora}\cite{Yang2016RevisitingSL}: A standard citation network benchmark where nodes represent machine learning papers and edges represent citations. The task is to classify each paper into one of seven predefined subject categories. Node features are bag-of-words representations of the paper's content.

    \item \textbf{Citeseer}\cite{Yang2016RevisitingSL}: Another widely-used citation network where nodes are scientific publications and edges are citations. The papers are classified into six categories. Similar to Cora, node features are based on the document's textual content.

    \item \textbf{CoraFull}\cite{McCallum2000AutomatingTC}: A larger citation network encompassing 70 classes. Nodes correspond to academic papers and edges denote citation links between them.

    \item \textbf{Arxiv}\cite{Hu2020OpenGB}: This dataset is constructed from the collaboration network of Computer Science (CS) papers on the arXiv platform, indexed by MAG \cite{Sinha2015AnOO}. Nodes, which represent CS papers, are classified into 40 subject areas. Node features are generated by averaging the word embeddings of the titles and abstracts.

    \item \textbf{Reddit}\cite{Hamilton2017InductiveRL}: A social network dataset derived from Reddit posts made in September 2014. Here, nodes are individual posts, and an edge connects two posts if the same user has commented on both. The objective is to classify posts into their corresponding communities (subreddits). Node features are engineered from post attributes such as title, content, score, and comment count.

    \item \textbf{Products}\cite{Hu2020OpenGB}: An Amazon product co-purchase network where nodes represent products and edges indicate that two products are frequently bought together. Node features are derived from dimensionality-reduced bag-of-words representations of the product descriptions.
\end{itemize}

\noindent 

\begin{figure*}[t!]
    \centering
    \includegraphics[width=1\linewidth]{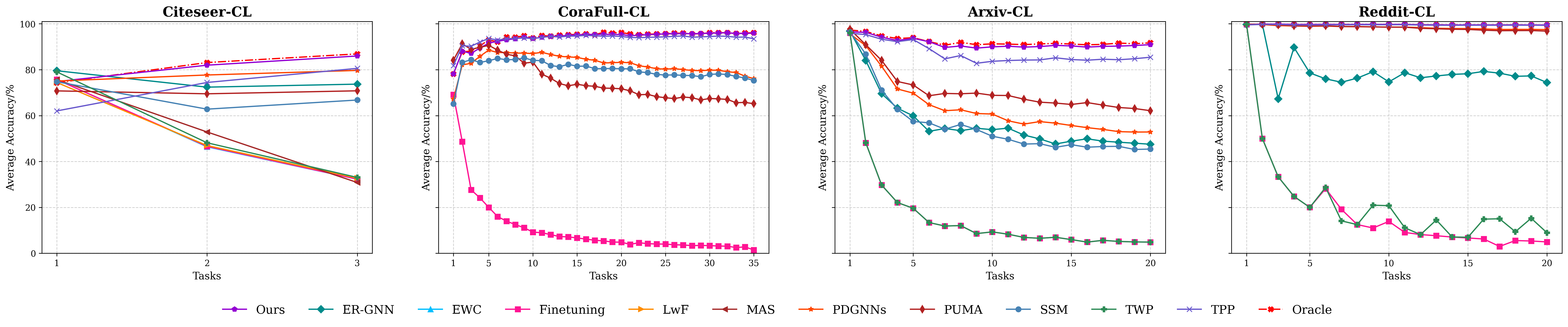}
      \caption{Performance comparison of our method, \textbf{TAAM} (Ours, purple line), against leading CGL methods on four datasets.}
  \label{fig:main_results}
\end{figure*}

\begin{figure*}[t!]
    \centering
    \includegraphics[width=0.8\linewidth]{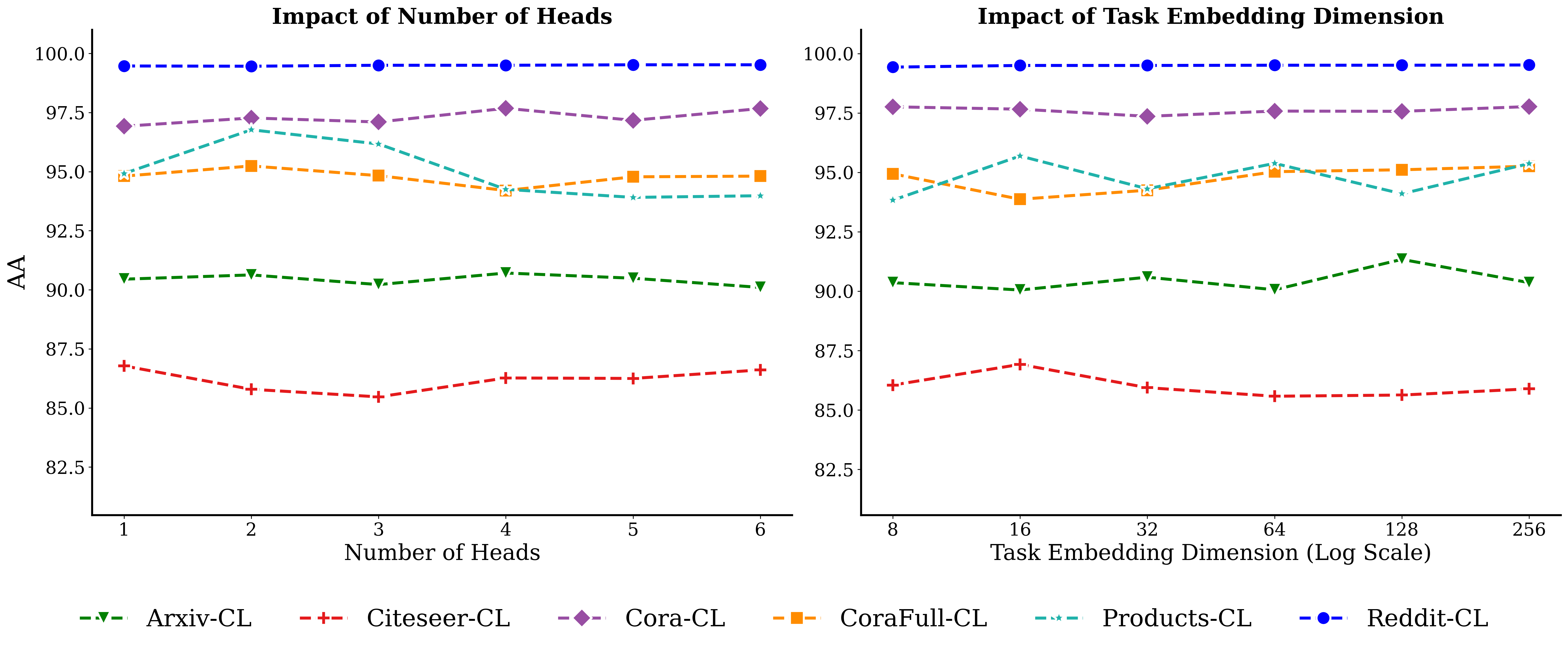}
      \caption{The impact of hyperparameters on TAAM varies across different datasets}
  \label{fig:hyperparam_sensitivity}
\end{figure*}

\section{D. Details of baselines}
We evaluate our method on a suite of leading CGL methods. including three mainstream categories based on Replay, regularization, and parameter isolation.

\begin{itemize}

    \item \textbf{Finetuning}\cite{gcn}is denotes the backbone GNN without continual learning technique. Therefore,this can be viewed as the lower bound on the continual learning performance.
        
    \item \textbf{EWC}\cite{ewc} is a regularization-centric approach that penalizes changes to parameters deemed crucial for previous tasks. This is achieved by adding a quadratic penalty proportional to the parameter's importance, thereby safeguarding performance on priorly learned tasks.
    
    \item \textbf{LwF}\cite{LwF} employs knowledge distillation to maintain performance on old tasks. It minimizes the discrepancy between the output logits of the previous and current models, effectively transferring knowledge without retaining old data.

    \item\textbf{MAS}\cite{mas} also preserves vital parameters from past tasks, but bases the importance on the sensitivity of the model's predictions to parameter shifts.

    \item\textbf{TWP} \cite{TWP} focuses on preserving important parameters within a topological aggregation framework and introduces a loss minimization strategy for previous tasks that relies on these preserved parameters.

    \item\textbf{ER-GNN} \cite{ERGNN} is a replay-based method that constructs a memory buffer by storing representative nodes selected from previous tasks.

    \item\textbf{SSM} \cite{SSM} enhances graph continual learning by explicitly incorporating topological information. It stores sparsified computation subgraphs of selected nodes in its memory.
    \item\textbf{PUMA} Based on CaT\cite{liu2023cat}, PUMA \cite{liu2024puma} enhances graph condensation coverage by incorporating both labeled and unlabeled nodes. It introduces a reinitialized optimization strategy that reconciles knowledge consolidation between historical and new graphs through balanced scratch training.

    \item\textbf{PDGNNs} \cite{Zhang2024Topology-awareembeddingmemoryforcl} decouple trainable parameters from computational subgraphs through topology-aware embeddings (TEs). These TEs are compressed representations of ego-subnetworks that significantly reduce memory overhead.

    \item\textbf{Joint} is base the backbone SGC can can get access to the data of all tasks, representing a theoretical upper bound of graph class-incremental learning
    
    \item\textbf{Oracle} is base the backbone SGC can can get access to the data of all tasksand task IDs, representing a stronger upper bound of graph class-incremental learning.
    
\end{itemize}

\section{E. Evaluation Metrics}

To comprehensively evaluate the model's performance in a continual learning setting, we adopt two standard metrics: Average Accuracy and Average Forgetting . Our evaluation protocol is based on the accuracy matrix $M \in \mathbb{R}^{T \times T}$, where $T$ represents the total number of tasks. An element $M_{t,j}$ of this matrix denotes the classification accuracy on task $j$ after the model has been sequentially trained up to task $t$.

\noindent\textbf{Average Accuracy (AA)} measures the overall performance of the model across all tasks after the entire training sequence is complete. It is calculated by averaging the accuracies on all tasks $j$ after the final task $T$ has been learned. A higher AA signifies a better-performing model. The formula is:
$$
\text{AA} = \frac{\sum_{j=1}^{T} M_{T,j}}{T}
$$

\noindent\textbf{Average Forgetting (AF)} quantifies how much the model forgets previously learned tasks as it acquires new knowledge. It is computed by averaging the difference between the peak accuracy on a task (i.e., immediately after it was learned) and its accuracy after training on a subsequent task, averaged over all past tasks. The formula is given by:
$$
\text{AF} = \frac{\sum_{j=1}^{T-1} (M_{T,j} - M_{j,j})}{T-1}
$$

\section{F. More Experiment Results}

\subsection{Performance Comparison}

As shown in Figure~\ref{fig:main_results}, the Figure illustrate the evolution of Average Accuracy (y-axis) as the number of tasks (x-axis) increases. TAAM consistently maintains high accuracy, closely tracking the \textbf{Oracle} performance (red line), which represents the theoretical upper bound. This is a direct result of our architectural design, which isolates and preserves past knowledge in dedicated, frozen Neural Sub-Modules (NSMs).

In stark contrast, the accuracy of competing methods visibly degrades. Regularization-based approaches like EWC and LwF suffer a severe performance collapse, while even advanced replay methods, including PDGNNs, exhibit a clear downward trend. By maintaining the integrity of learned knowledge while adapting to new tasks, TAAM presents a more effective solution to the fundamental stability-plasticity dilemma.

\subsection{Hyperparameter Analysis}
We investigated the sensitivity of our model to two key hyperparameters: the number of attention heads ($K$) and the task embedding dimension ($d_e$). Figure~\ref{fig:hyperparam_sensitivity} plots the Average Accuracy (AA) against variations in these parameters.

\noindent\textbf{Number of Heads.} The left panel of Figure~\ref{fig:hyperparam_sensitivity} shows that model performance is largely robust to the number of heads. However, a modest improvement is observed on several datasets (e.g., Cora-CL, CoraFull-CL) when increasing the number of heads from one to two or three, after which performance tends to plateau. This suggests that while a single head is effective, a multi-head configuration can capture more complex, task-specific feature modulations, confirming the utility of the attention mechanism.

\noindent\textbf{Task Embedding Dimension.} As shown in the right panel, the model is remarkably insensitive to the task embedding dimension. Performance remains stable across a wide range, from 8 to 256. Notably, a highly compact embedding (e.g., $d_e=16$) achieves results on par with a much larger one. This finding underscores the parameter efficiency of our approach, as it can generate effective, specialized modulators from very low-dimensional task representations.

% \bibliography{aaai2026}

% \end{document} 
\end{document}